\documentclass[lettersize,journal]{IEEEtran}
\usepackage{amsmath,amsfonts}
\usepackage{algorithmic}
\usepackage{algorithm}
\usepackage{array}
\usepackage[caption=false,font=normalsize,labelfont=sf,textfont=sf]{subfig}
\usepackage{textcomp}
\usepackage{stfloats}
\usepackage{url}
\usepackage{verbatim}
\usepackage{graphicx}
\usepackage{cite}
\usepackage{multirow}% new added
\begin{document}

\title{AlignVE: Visual Entailment Recognition \\Based on Alignment Relations}

\author{Biwei~Cao,
        Jiuxin~Cao,
        Jie~Gui,
        Jiayun~Shen,
        Bo~Liu,
        Lei~He,
        Yuan~Yan~Tang,
        and~James~Tin-Yau~Kwok
        
        % <-this % stops a space

\IEEEcompsocitemizethanks{
\IEEEcompsocthanksitem This work was supported by National Key R\&D Project of China under Grant No.2021QY2102; National Natural Science Foundation of China under Grants No.62172089, No.61972087, No.62172090, No.62106045, No.62172458; Natural Science Foundation of Jiangsu Province under Grant No.BK20191258; Jiangsu Provincial Key Laboratory of Computer Networking Technology; Jiangsu Provincial Key Laboratory of Network and Information Security under Grant No. BM2003201; Key Laboratory of Computer Network and Information Integration of Ministry of Education of China under Grant No. 93K-9; CAAI-Huawei MindSpore Open Fund; Alibaba Group through Alibaba Innovative Research Program.\protect
\IEEEcompsocthanksitem Biwei Cao, Jiuxin Cao, Jie Gui and Jiayun Shen are with the School of Cyber Science and Engineering, Southeast University, Nanjing 211189, China, and also with the Key Laboratory of Computer Network and Information of Ministry of Education of China, Nanjing 211189, China. Jiuxin Cao and Jie Gui are also with Purple Mountain Laboratories, Nanjing 210000, China.\protect\\
E-mail: \{caobiwei, jx.cao, guijie, jyshen\}@seu.edu.cn.
\IEEEcompsocthanksitem Bo~Liu is with the School of Computer Science and Engineering, Southeast University, Nanjing 211189, China.\protect\\
E-mail: bliu@seu.edu.cn.
\IEEEcompsocthanksitem Lei~He is with Information Engineering University, China and Purple Mountain Laboratories, Nanjing 210000, China.\protect\\
E-mail: helei@pmlabs.com.cn.
\IEEEcompsocthanksitem Yuan Yan Tang is with the Department of Computer and Information Science, University of Macao, Macao 999078, China.\protect\\
E-mail: yuanyant@gmail.com.
\IEEEcompsocthanksitem James Tin-Yau Kwok is with the Department of Computer Science and Engineering, The Hong Kong University of Science and Technology, Hong Kong 999077, China.\protect\\
E-mail: jamesk@cse.ust.hk.
}% <-this % stops an unwanted space
\thanks{(Corresponding author: Jiuxin Cao, Jie Gui)}}
        
        % <-this % stops a space



\maketitle

\begin{abstract}
 Visual entailment (VE) is to recognize whether the semantics of a hypothesis text can be inferred from the given premise image, which is one special task among recent emerged vision and language understanding tasks. Currently, most of the existing VE approaches are derived from the methods of visual question answering. They recognize visual entailment by quantifying the similarity between the hypothesis and premise in the content semantic features from multi modalities. Such approaches, however, ignore the VE's unique nature of relation inference between the premise and hypothesis. Therefore, in this paper, a new architecture called AlignVE is proposed to solve the visual entailment problem with a relation interaction method. It models the relation between the premise and hypothesis as an alignment matrix. Then it introduces a pooling operation to get feature vectors with a fixed size. Finally, it goes through the fully-connected layer and normalization layer to complete the classification. Experiments show that our alignment-based architecture reaches 72.45\% accuracy on SNLI-VE dataset, outperforming previous content-based models under the same settings. 
\end{abstract}

\begin{IEEEkeywords}
Computer vision, visual entailment, alignment relation.
\end{IEEEkeywords}

\section{Introduction}

\IEEEPARstart{V}{isual} entailment (VE) proposed by Xie et al. \cite{1,2} is a multi-modal inference task derived from the original single-modal textual entailment (TE) \cite{3} in natural language processing research and visual question answering (VQA) \cite{4}. In entailment recognition, given a premise $P$ and a hypothesis $H$, the system outputs the \textit{entailment} if $H$ can be concluded from $P$. The result is a \textit{contradiction} if $H$ contradicts  $P$ and otherwise it is \textit{neutral}, which means $P$ and $H$ are not related. For example, in Table~\ref{table0}, when the hypothesis text is ``Two women are holding packages.", the entity of ``two women" and the action of ``holding packages" are able to be matched with the contents of the premise image. Therefore, the hypothesis is included in premise and the entailment relation result is \textit{entailment}. In the hypothesis sentence ``The sisters are hugging goodbye while holding to go packages after just eating lunch.", the conclusion of ``sisters" is neither able to be drawn from this premise image nor be verified to contradict the image content. The entailment relation of this pair of premise and hypothesis is thus \textit{neutral}. When the hypothesis becomes ``The men are fighting outside a deli.", the verb ``fighting" is contradicted with the action of the figures in the premise image, so that the result of the entailment relation is \textit{contradiction}.
\begin{table*}[t]
\renewcommand\arraystretch{1}
\centering

\caption{A Visual Entailment Example with a premise image and three kinds of entailment relation.}
\begin{tabular}{l|l|l}\hline
     \multirow{10}*{\includegraphics[width=2.5in,height=1.2in,clip,keepaspectratio]{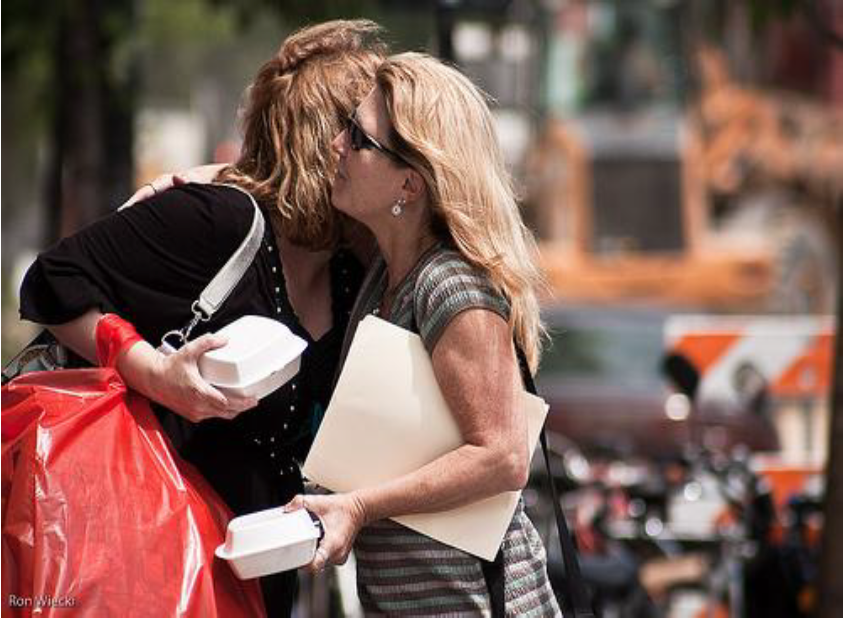}}& & \\
     & Two women are holding packages. & \textit{Entailment}\\
     &&\\
     \cline{2-3}
     &&\\
     & The sisters are hugging goodbye & \multirow{2}*{\textit{Neutral}}\\
     & while holding to go packages after just eating lunch.&\\
     &&\\
     \cline{2-3}
     &&\\
     & The men are fighting outside a deli.& \textit{Contradiction}\\
     &&\\\hline
     Premise & Hypothesis & Entailment Relation\\\hline
\end{tabular}

\label{table0}
\end{table*}

Recently, the majority of VE studies \cite{1,2,5} are inspired by VQA. It is found that VQA includes a relative wide range of specific tasks, such as generating a sentence as an answer \cite{6,7}, discriminating yes/no question \cite{4,8}, selecting a word or phrase in predefined answers \cite{4} and filling in the blank \cite{9}. What is more, the answers in VQA datasets \cite{4} mainly consist of objects, colors, categories, scenes, sports and even brands. Thus, it can be concluded that VQA is a broader multi-modal task compared to VE. Because of this, most of the VQA methods are universal multi-modal classification models, which extract and fuse multi-modal features to build enriched content semantic features. While for VE task, its goal is to only classify the relation between the premise and hypothesis into three classes. Although the recent VQA methods are universal and can be easily applied to VE, they are not designed specifically for VE. The works which solve VE tasks by simply adopting VQA methods ignore the characteristic of VE — recognizing relations. On the basis of this observation, we believe that modeling the relation between premise and hypothesis should be meaningful and essential to VE studies.

The main contributions of our work can be summarised as follows. 

Firstly, based on the difference between a VQA task and a VE task, an alignment-relation-based visual entailment architecture AlignVE is proposed, particularly considering the characteristics of the VE task. It contains three parts, namely visual feature extraction module, text feature extraction module and the alignment-based classifier construction module. The first two modules aim to extract visual and text features from premise and hypothesis. The alignment-based classifier construction module attempts to build the relation interaction between these features and utilize the obtained alignment relation to complete the classification task.

Secondly, the relation alignment method is firstly introduced in this field for the enhancement of recognizing the interaction between the premise image and hypothesise text, specifically modeling the entailment relation for VE classification. 

Last but not least, comprehensive experiments are conducted with SNLI-VE dataset to re-evaluate existing VE models and evaluate both migrated TE models and our proposed model. The results indicate that our AlignVE architecture outperforms the previous methods and still keeps the simplicity.

\section{Related Work}
\IEEEpubidadjcol
Text entailment (TE), a classic natural language processing task, is the predecessor task of VE and has developed greatly in entailment recognition. Derived from TE, VE is defined as a multi-modal classification task, which is closely related to VQA, a typical multi-modal classification task that has been extensively studied in recent years. The VQA study history reflects the development of multi-modal classification methods. Both of TE and VQA have a close relation and inspiration to VE.
\subsection{Textual Entailment}
TE is proposed as the PASCAL recognizing textual entailment challenge \cite{3} in 2005. In early studies, most of them are based on similarity measurement \cite{11,12,13}, alignment algorithms \cite{14,15}, machine learning \cite{16} and logic inference \cite{17} methods.

Since the large dataset SNLI \cite{18} was published, deep-learning based methods are used in the majority of recent studies. RNN like LSTM is applied to TE \cite{18}, combined with attention mechanism \cite{19,20}. More LSTM-based methods have been developed like match-LSTM  (mLSTM) \cite{21}, re-read LSTM (rLSTM) \cite{22} and Enhanced Sequential Inference Model (ESIM) \cite{23}. CNN-based methods have also been developed for TE, e.g. TBCNN (Tree-based CNN) \cite{24,25}. The Interactive Inference Network (IIN) \cite{26} explicitly models the relation between the premise and hypothesis as an interaction space and utilizes a CNN to extract features in it. The logic-based methods also have progresses recently \cite{27}. The TE studies analyze the entailment recognition task and propose specific methods designed for this, which inspire VE studies.

\subsection{Visual Question Answering}
The VQA is formally defined as providing an accurate natural language answer to a given image and a natural language question about the image \cite{4}. Early studies \cite{7,27} recognized VQA as a natural language generation task and then handled it with the encoder-decoder framework. However, recent studies mostly model VQA as a classification task which selects an answer from a predefined corpus. The typical paradigm of classification approach is that the VQA system extracts image features with CNN or object detector, then encodes the question with RNN and finally implements several mechanisms to do feature fusion and classification. To improve the quality of multi-modal feature extraction and fusion, many methods based on the attention mechanism have been proposed for VQA, such as region selection attention \cite{29}, co-attention \cite{30}, stacked attention \cite{31}, high-order attention \cite{32}, bottom-up and top-down attention \cite{33}, dense co-attention \cite{34}, modular co-attention \cite{35}, dynamic fusion with intra-and inter-modality attention \cite{36} and multi-grained attention \cite{37}. In conclusion, the design of VQA methods becomes more complicated together with the trend of fine-grained, multi-modal and multi-hop attention, thus enriching the representation of image and question features. 
%2022-07-01
Also, the model CLIP \cite{50} utilizes the natural language supervision information as the training signal to learn the visual feature, and takes the dot product method to build the interaction between image and text domain in multi-modal embedding space. These methods are analyses of images and texts at the macro level, and lack of a fine-grained analysis into the entailment relation between images and texts. 

\subsection{Visual Entailment}
The current number of VE studies is limited \cite{1,2,5,38}. Most methods are deep-learning based. Xie et al. \cite{1,2} propose the Explainable VE (EVE) architecture, which is based on Bottom-Up/Top-down architecture \cite{33}, the first place of the 2017 VQA challenge. Their intuition is that both of VE and VQA can be modeled as multi-modal classification tasks. By modifying typical VQA models for VE task, EVE can benefit from proved multi-modalilty feature fusion and high classification performance of VQA models. Do et al. \cite{5} re-annotate SNLI-VE dataset, re-evaluate existing models and introduce a new task e-SNLI-VE 2.0 that requires explanation sentence generation for VE recognition. Chen et al. \cite{39} propose a universal pretrained image-text representation model UNITER and include VE as one of their downstream tasks.

Besides, Suzuki et al. \cite{38} propose a logic-based VE system. This system translates an image into a scene graph or first-order logic (FOL) structure and at the same time, parses sentences into FOL formulae. Then, this system applies Prover9\footnote{http://www.cs.unm.edu/mccune/prover9/} as the inference engine to conduct theorem proofs. However, this method is slow and can only recognize VE as a binary classification task since it relies on a timeout (10s) to decide whether a hypothesis is entailed or not.

Therefore, we can draw a conclusion for the relationship among TE, VQA and VE in Table~\ref{tab00}. 

In all, most VE studies lack analysis on the characteristics of VE, thus lacking the study on the interaction relation between the premise and hypothesis. New methods are needed to fill up this gap. Therefore, our AlignVE architecture is designed to model the premise-hypothesis relation as an alignment matrix and recognizes VE based on it. It is designed specifically for VE and maintains the simplicity of the whole structure. To the best of our knowledge, there is no relation-based VE study currently. It can be viewed as a significant starting point of relation-based VE methods.

\begin{table*}[t]
\renewcommand\arraystretch{1}
\centering
%\resizebox{.95\columnwidth}{!}{
\caption{Relationship Among TE, VQA and VE in their goals, result categories and methods.}
\begin{tabular}{l|c|c|c}\hline
    & TE & VQA & VE\\\hline
    \multirow{2}*{Goal} & \multirow{2}*{Recognize the relationship between sentences.}   & \multirow{2}*{Answer visual questions.}  & Recognize the relationship\\
     &  &  & between the image and sentence.\\\hline
    \multirow{2}*{Category} &is entailment or not (early stage), &Yes/No, object, color, number,  & \multirow{2}*{entailment / neutral / contradiction}\\
    &entailment / neutral / contradiction& position, multiple choices...&\\\hline
    \multirow{2}*{Method} & manually crafted features (early stage),& deep learning generation,  & deep learning classification, \\
     & logic inference, deep learning classification &deep learning classification&logic inference\\\hline

\end{tabular}

\label{tab00}
\end{table*}
\section{The AlignVE Architecture}
\begin{figure}[t]
\centering
\includegraphics[width=1\columnwidth]{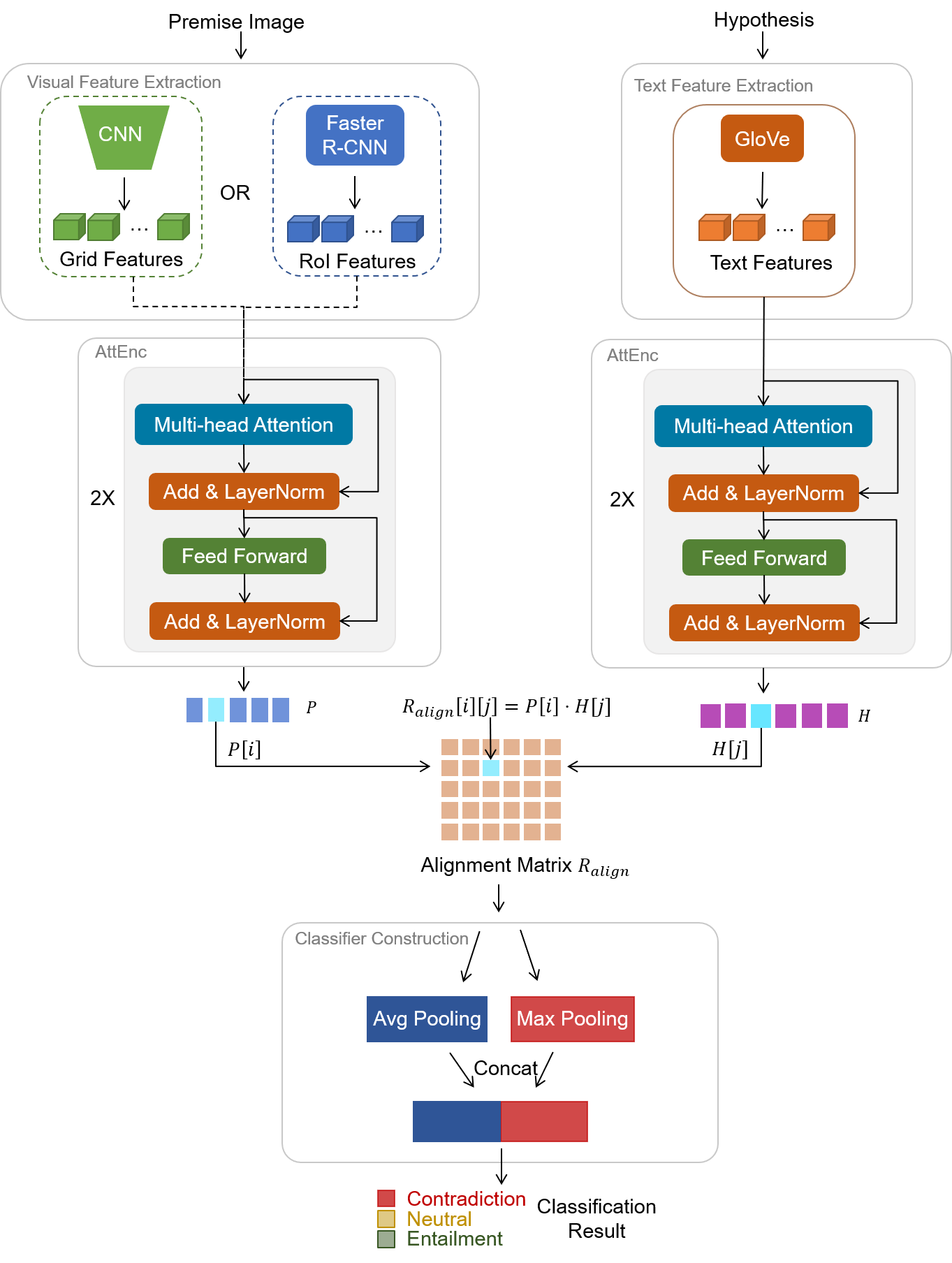} 
\caption{The Architecture of AlignVE. The main components of the AlignVE are visual extraction module, text feature extraction module and alignment-based classifier construction module.}
\label{fig1}
\end{figure}
The VE task is defined as a multi-modal task which recognizes the relation between a premise image and hypothesis text. Formally, given a premise image $P_{image}$ and a hypothesis text $H_{text}$, the goal of VE is to determine if $H_{text}$ can be concluded from $P_{image}$ and to classify this relation into three classes: \textit{Entailment}, \textit{Contradiction} and \textit{Neutral}.

\textit{Entailment} holds if the evidence in $P_{image}$ is enough to conclude that $H_{text}$ is true.

\textit{Contradiction} holds if the evidence in $P_{image}$ is enough to conclude that $H_{text}$ is false.

\textit{Neutral} holds otherwise, implying the evidence in $P_{image}$ is insufficient to draw a conclusion about $H_{text}$.

The overall architecture of AlignVE is shown in Figure \ref{fig1} which mainly contains three modules as follows.

\begin{itemize}
\item  
The visual feature extraction module, which uses a CNN to extract grid features or an object detector to extract RoI features, and then encodes the visual features with an AttEnc.
\item  
The text feature extraction module, which uses GloVe \cite{10} as word embedding and encodes the text features with AttEnc. 
\item  
The alignment-based classifier construction module, which is to calculate the alignment matrix between the premise and hypothesis, and then apply adaptive pooling to get a fixed-length feature vector for classification.
\end{itemize}

\subsection{Visual Feature Extraction Module}
The visual feature extraction module takes an image $P_{image}$ as the input premise and outputs a sequence of vectors $P$ to represent the premise image. Two approaches are taken to extract image features. One inputs images to a pretrained CNN and takes the feature map as image grid features, and the other one inputs images to a pretrained object detector (Faster R-CNN) and takes the RoI features as image features. Both approaches are widely applied in visual understanding fields and several related works \cite{33,40} make a detailed comparison and analysis between the grid and RoI features. This process can be represented as 
\begin{equation}
    F_p=CNN(P_{image})\tag{1}
\end{equation}
or
\begin{equation}
    F_p=RCNN(P_{image}).\tag{2}
\end{equation}

In our study, $F_{p}\in \mathbb{R}^{m\times d_{p}}$ where m is fixed to 36 for both gird or RoI features and $d_{p}$ is 2048 according to the design of ResNet \cite{42}.

Then, the extracted sequence of visual features is encoded by a self-attention encoder $AttEnc$ from Transformer \cite{41},
\begin{equation}
    P=AttEnc(F_p).\tag{3}
\end{equation}
Here, the dimensionality of each feature vector in $P\in \mathbb{R}^{m\times d}$ is transformed to a unified feature dimensionality $d$.
\subsubsection{Grid Features}
Specifically, for image grid features, a pretrained ResNet101 \cite{42} is used as CNN and the output feature map from last layer of conv5\_x before pooling is taken as grid features. Due to the varying spatial sizes of different images, the output feature maps are in different sizes. The bilinear interpolation is selected to resize the feature map with variable spatial size to the one of fixed size $6\times6$.
\subsubsection{RoI Features}
For image RoI features, a pretrained Faster R-CNN \cite{43} is taken as object detector to detect salient objects and RoI features  $F_{p}\in \mathbb{R}^{m\times d_{p}}$ is obtained by mapping RoI boxes on the feature map of backbone with RoIAlign. Top-36 RoI features are selected by ranking with the RoI’s bounding box score. For the cases of an image containing RoIs less than 36, zero-vector is padded to it.

\subsubsection{Self-Attention Encoder}
To achieve better representation of an image, the self-attention encoder proposed by Vaswani et al. \cite{41} in Transformer architecture is applied instead of using the grid/RoI features directly. We call this AttEnc for short. AttEnc is powerful and faster than RNN like LSTM \cite{44} or GRU \cite{45} because its calculation flow is designed in a parallel way, not using iterations. It is applied to enrich the image features with their context information and is responsible to transform the input dimensionality to a unified feature dimensionality $d$.

For an AttEnc, it takes a sequence of features $F_p$ as input: 

\begin{equation}
    F_{enc}=AttEnc(F_p),\tag{4}
\end{equation}
where $F_p\in \mathbb{R}^{m\times d_{p}}$.

In a basic building block of AttEnc, the scale-dot product (SDP) attention \cite{41} is taken for the input features,
\begin{equation}
    Att_{SDP}(Q,K,V)=softmax(\frac{QK^T}{\sqrt{d_p}})V.\tag{5}
\end{equation}

Based on SDP attention, the multi-head attention \cite{41} is designed to pay attention to information from different representation subspaces at different positions, 

\begin{equation}
    Att_{MH}(Q,K,V)=concat(head_1,\dots,head_h)W^O,\tag{6}
\end{equation}
where $W^O$ is the projection parameter matrix.
In self-attention mechanism, the query $Q$, key $K$ and value $V$ are input features with different linear transformations as (\ref{eq7}). For the $i$-th head, it comes from an SDP attention in a representation subspace,

\begin{equation}
\label{eq7}
    head_i=Att_{SDP}(F_pW_i^Q,F_pW_i^K,F_pW_i^V),\tag{7}
\end{equation}
where $W_i^Q$, $W_i^K$ and $W_i^V$ are projection parameter metrics. 
The attended features $F_{att} \in \mathbb{R}^{m\times d_p}$, which are obtained from the self-attention mechanism, are then passed to a feed-forward network with residual connection \cite{42} and layer normalization \cite{46} as 
\begin{equation}
\label{eq8}
    F_{enc}=LN(ReLU(LN(F_{att}+F_p)W_f+b_f)+F_{att}).\tag{8}
\end{equation}
The output of AttEnc is $F_{enc}\in \mathbb{R}^{m\times d}$, where $m$ is the length of input sequence of features and $d$ is the target feature dimensionality.

\subsection{Text Feature Extraction Module}
The text feature extraction module takes a sequence of text $T$ as the input hypothesis and outputs a sequence of vectors $H$ to represent the hypothesis text. At the beginning, the text is tokenized and the GloVe \cite{10} is used as word embedding to represent each token,
\begin{equation}
    F_{h}=GloVe(T).\tag{9}
\end{equation}

For an $n$-token text, the primary representation sequence $F_h\in \mathbb{R}^{n\times d_h}$ means each token of the text is represented by a vector whose size is $d_h$.

Next, with the aim of injecting the relative position information into the text sequence, the positional encoding $PE$ is added to text,
\begin{equation}
    F_{h}=F_h+PE.\tag{10}
\end{equation}

Specifically, the position encoding is referenced from the sine and cosine functions in Transformer \cite{41},
\begin{equation}
    PE_{(pos,2i)}=\sin(\frac{pos}{10000^{\frac{2i}{d}}}),\tag{11}
\end{equation}
\begin{equation}
    PE_{(pos,2i+1)}=\cos(\frac{pos}{10000^{\frac{2i}{d}}}),\tag{12}
\end{equation}
where $pos$ is the position and $i$ is the dimension. 
In the end, the self-attention encoder is utilized to encode the word representations with their context information and transform the input dimensionality to a unified feature dimensionality $d$,
\begin{equation}
    H=AttEnc(F_h).\tag{13}
\end{equation}
Here, the $H\in \mathbb{R}^{n\times d}$ is transformed in a unified feature dimensionality $d$.

\subsection{Alignment Relations}
The alignment relations are modeled as an alignment matrix $R_{align}$.
Given the premise image features $P$ from the visual feature extraction module and hypothesis text features $H$ from the text feature extraction module, the alignment value of the $i$-th premise vector $P[i]$ and the $j$-th hypothesis vector $H[j]$ is the dot-product of both $d$-dim vector,
\begin{equation}
    R_{align}[i][j]=P[i]\cdot H[j],\tag{14}
\end{equation}
where $\cdot$ represents vector dot-product.

The alignment matrix can be written in a matrix computation way,

\begin{equation}
    R_{align}=PH^T,\tag{15}
\end{equation}
where $R_{align}\in \mathbb{R}^{m\times n}$ is the alignment matrix representing the alignment relations between $m$ premise image features and $n$ hypothesis text features.

\subsection{Classifier Construction}

Considering the shape of the alignment matrix $m\times n$ is variable for different premise-hypothesis pairs, it has to be down-sampled into a fixed size for the further classification task. Therefore, we choose to use the adaptive pooling method, which is able to obtain a specified output size no matter what the size of the input is, on this 2-D matrix. Here, we use the concatenation of adaptive average pooling and adaptive max pooling, with the purpose of enriching the feature information. To be more specific, given the 2-D alignment matrix, we do the adaptive average pooling and max pooling operations concurrently to achieve two fixed-size feature vectors respectively. Each pooling method pools the alignment matrix to a 150-D vector, and then the two pooled 150-D vectors are concatenated to form a 300-D feature vector, which is used as the input of the fully-connected layer for the classification. This process is formulated in Eq. (\ref{A}) - (\ref{D}).

\begin{equation}
    V_{avg}=avgpool(R_{align}),\tag{16}
\label{A}
\end{equation}
\begin{equation}
    V_{max}=maxpool(R_{align}),\tag{17}
\label{B}
\end{equation}
\begin{equation}
    V=concat(V_{avg},V_{max}),\tag{18}
\label{C}
\end{equation}
\begin{equation}
    S=softmax(VW_v+b_v),\tag{19}
\label{D}
\end{equation}
where $avgpool$ and $maxpool$ are adaptive average pooling and adaptive max pooling respectively. $S\in \mathbb{R}^C $is the predicted score for VE classification, $C$ is the number of classes and both $W_v$ and $b_v$ are parameters learned through training process.

\section{Experiments}
\begin{table*}[t]
\renewcommand\arraystretch{1}
\centering
\setlength{\tabcolsep}{5.15mm}

\caption{SNLI-VE dataset distribution in image and three kinds of entailment relations as well as the splitting way.}
\begin{tabular}{l|l|l|l|l}\hline
     & Training  & Validation& Testing & Total \\\hline
    Image &  29,783  &  1,000 & 1,000&\textbf{ 31,783} \\
    \#Entailment& 176,932 & 5,959 & 5,973 & 188,864 \\
    \#Neutral   & 176,045 & 5,960 & 5,964 & 187,969 \\
    \#Contradiction & 176,550 & 5,939 & 5,964 & 188,453\\
    \#Total        & 529,527 & 17,858 & 17,901 & \textbf{565,286}\\\hline

\end{tabular}

\label{table3}
\end{table*}

\begin{table*}[t]
\renewcommand\arraystretch{1}
\centering
\setlength{\tabcolsep}{5.15mm}

\caption{SNLI-VE-2.0 dataset distribution in image and three kinds of entailment relations as well as the splitting way.}
\begin{tabular}{l|l|l|l|l}\hline
     & Training  & Validation& Testing & Total \\\hline
    Image &  29,783  &  1,000 & 1,000&\textbf{ 31,783} \\
    \#Entailment& 131,023 & 5,254 & 5,218 & 141,495 \\
    \#Neutral   & 125,902 & 3,442 & 3,801 & 133,145 \\
    \#Contradiction & 144,792 & 5,643 & 5,721 & 156,156\\
    \#Total        & 401,717 & 14,339 & 14,740 & \textbf{430,796}\\\hline

\end{tabular}

\label{table003}
\end{table*}

\begin{table*}[t]
\renewcommand\arraystretch{1}
\centering

\caption{Experiment results on SNLI-VE dataset provided by previous papers \cite{1,2} and this paper's implementation, including the re-implemented models, transferred models and our proposed model AlignVE. }
\begin{tabular}{l|l|c|c}\hline
    Model Type & Model & Validation Accuracy (\%) & Test Accuracy (\%) \\\hline
    \multirow{4}*{Original Model} &TD         & 70.53 & 70.30 \\
    &BUTD       & 69.34 & 68.90 \\
    &EVE-Image  & 71.56 & 71.16 \\
    &EVE-RoI    & 70.81 & 70.47 \\
    % \hline\hline
    % Do et al.:  &      &      \\
    % BUTD       & -     & 73.02 \\
    \hline \hline

     \multirow{4}*{Re-implemented Model}& TD         & 71.00 & 71.04 \\
    &BUTD       & 70.52 & 70.97 \\
    &EVE-Image  & 71.52 & 71.43 \\
    &EVE-RoI    & 70.43 & 70.43 \\
    \hline
     \multirow{4}*{Transferred Model}&rLSTM-Grid & 70.92 & 71.25 \\
    &rLSTM-RoI  & 71.24 & 71.26 \\
   & IIN-Grid   & 71.02 & 71.34 \\
   & IIN-RoI    & 71.33 & 71.30 \\
   \hline
   \multirow{2}*{Our Proposed Model} & AlignVE-Grid & \textbf{72.31} & \textbf{72.45} \\
   & AlignVE-RoI& 72.02 & 72.20 \\\hline
\end{tabular}

\label{table1}
\end{table*}

\begin{table*}[t]
\renewcommand\arraystretch{1}
\centering
\setlength{\tabcolsep}{6mm}

\caption{Test accuray (\%) of BUTD model and our proposed model AlignVE on SNLI-VE and SNLI-VE-2.0 dataset.}
\begin{tabular}{l|c|c}\hline
    Model & SNLI-VE dataset & SNLI-VE-2.0 dataset \\\hline

    BUTD       & 70.97 & 69.26\\
    AlignVE-Grid & 72.45 & 72.67\\\hline
\end{tabular}

\label{tab01}
\end{table*}
\begin{table*}[t]
\renewcommand\arraystretch{1}
\centering
\setlength{\tabcolsep}{6mm}
\caption{Ablation experiment results of AttEnc with MLP and GRU in accuracy (\%) on SNLI-VE dataset.}

\begin{tabular}{l|c|c|c|c}\hline
    Model & 
    \multicolumn{2}{c|}{Before ablation} 
    &\multicolumn{2}{c}{After ablation} \\\hline
    (replace AttEnc)& Validation & Test & Validation & Test \\\hline
    rLSTM-Grid & 70.92 & 71.25 & 69.65 & 69.34\\
    rLSTM-RoI  & 71.24 & 71.26 & 69.73 & 70.07\\
    IIN-Grid   & 71.02 & 71.34 & 69.99 & 70.30\\
    IIN-RoI    & 71.33 & 71.30 & 69.55 & 70.11\\\hline
    AlignVE-Grid & \textbf{72.31} & \textbf{72.45} & \textbf{71.25} & 70.63\\
    AlignVE-RoI& 72.02 & 72.20 & 70.73 & \textbf{71.01}\\\hline
    
\end{tabular}

\label{table2}
\end{table*}

\begin{table*}[t]
\renewcommand\arraystretch{1}
\centering
\setlength{\tabcolsep}{6mm}
\caption{Ablation experiment results of Alignment Matrix with a co-attention layer in accuracy (\%) on SNLI-VE dataset.}
% \resizebox{.95\columnwidth}{!}{
\begin{tabular}{l|c|c|c|c}\hline
    Model & 
    \multicolumn{2}{c|}{Before ablation} 
    &\multicolumn{2}{c}{After ablation} \\\hline
    (replace Alignment Matrix)& Validation & Test & Validation & Test \\\hline
    AlignVE-Grid & \textbf{72.31} & \textbf{72.45} & 66.3 & 66.01\\
    AlignVE-RoI& 72.02 & 72.20 & \textbf{66.64 }& \textbf{66.78}\\\hline
    
\end{tabular}

\label{tableCo}
\end{table*}

\begin{table}[t]
\renewcommand\arraystretch{1}
\centering
\setlength{\tabcolsep}{6mm}
\caption{ResNet series experiment results of the AlignVE-Grid model in accuracy (\%) on SNLI-VE dataset.}
% \resizebox{.95\columnwidth}{!}{
\begin{tabular}{l|c|c}\hline
    Model & Validation & Test\\\hline
   ResNet-18 & 70.27 & 70.25 \\
    ResNet-34 & 70.49 & 70.48\\
    ResNet-50   & 70.46 & 70.79 \\
    ResNet-101   & \textbf{72.31}& \textbf{72.45} \\
    ResNet-152 & 71.02 & 71.3 \\\hline
\end{tabular}

\label{tableResNet}
\end{table}

\begin{figure*}[t]
\centering
\includegraphics[width=2\columnwidth]{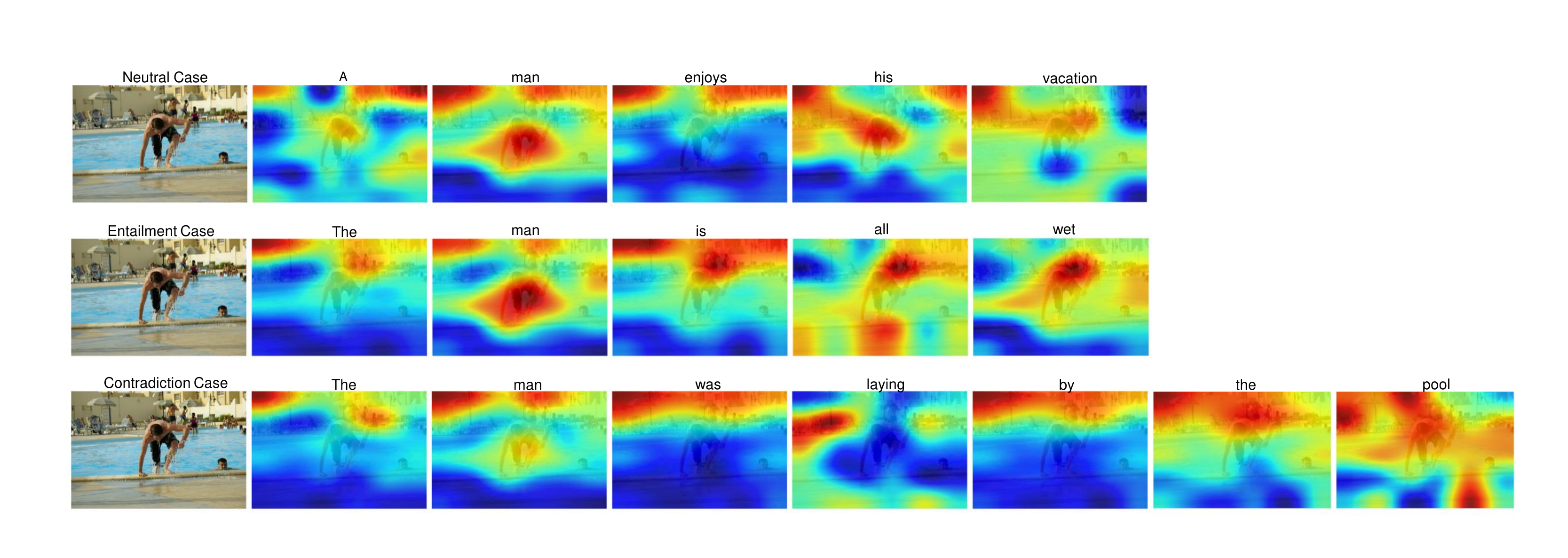} % Reduce the figure size so that it is slightly narrower than the column. Don't use precise values for figure width.This setup will avoid overfull boxes.
\caption{Alignment matrix visualization of AlignVE in the ``\textit{neutral}", ``\textit{entailment}" and ``\textit{contradiction}" cases. Premise images are on the leftmost part and the visual attention images according to each word of the hypothesises are on the right. In a visual attention image, the color from blue to red represents the alignment relation from weak to strong.}.
\label{fig2}
\end{figure*}

\subsection{Dataset}
Models are evaluated on SNLI-VE\footnote{https://github.com/necla-ml/SNLI-VE}  \cite{1,2} dataset and SNLI-VE-2.0\footnote{https://github.com/maximek3/e-ViL} \cite{5}. The SNLI-VE dataset proposed by Xie et al. is a VE dataset created by finding the corresponding image in Flickr30K dataset for each text premise-hypothesis pair in SNLI through the annotation of SNLI dataset. As shown in Table~\ref{table3}, 31,783 images and 565,286 premise-hypothesis pairs are labeled for 3 classes as \textit{entailment}, \textit{neutral} and \textit{contradiction}. We use the same dataset SNLI-VE published by Xie et al. for fair comparison. The SNLI-VE-2.0 dataset is built by Do et al. based on the SNLI-VE dataset with the improvement on the classification accuracy by re-annotating the incorrect data in \textit{neutral} class of validation set and test set. This dataset distribution is shown in Table~\ref{table003}, with 430,796 premise-hypothesis pairs for the 3-class classification. Do et al. \cite{5} mention that they try to see the difference in performance on the corrected validation and test sets by reproduction of BUTD model. However, the statistics of Do et al.'s work \cite{5} shows that the corrected dataset has no effect on the BUTD model performance by a drop of test accuracy from SNLI-VE dataset to SNLI-VE-2.0 dataset. Since Do et al. have completed related experiments using SNLI-VE-2.0, we also include this dataset and implement models on it for a better comparison.

\subsection{Baselines}
We select several models in recent VE studies \cite{1,2,5} as baseline models and migrate several TE architectures \cite{22,26} to the VE task for a better comparison.

\subsubsection{TD/BUTD}
This Bottom-Up Top-Down (BUTD) attention architecture is originally proposed by Anderson et al. \cite{33} for VQA and image caption. The ‘bottom-up’ attention stands for RoI feature from object detection in practice since the object detection can be considered as a hard attention on images. For convenience, the grid-feature-based Top-Down attention is named as TD and the RoI feature based Bottom-Up Top-Down attention as BUTD. Xie et al. \cite{1,2} evaluate both TD and BUTD as baseline models. Do et al. \cite{5} re-evaluate BUTD only. In our experiment, we implement this architecture as one of the baselines and re-evaluated both TD and BUTD in the experiment environment.

\subsubsection{EVE}
The Explainable Visual Entailment (EVE) model is a VE architecture proposed by Xie et al. \cite{1} by modifying the BUTD Attention model. Two branches are composed of EVE architecture: the EVE-Image and EVE-RoI. The EVE-Image takes grid features as image features while the EVE-RoI takes RoI features as image features. Both EVE-Image and EVE-RoI are re-evaluated by using our experiment settings.

\subsubsection{rLSTM}
The re-read LSTM (rLSTM) architecture is originally proposed by Sha et al. \cite{22} as a LSTM model specifically designed for TE. The rLSTM architecture is designed to enrich the content feature of hypothesises, by re-reading the premise while reading the hypothesis in LSTM iterations. In experiment, we migrate the rLSTM to VE by replacing the original text-only input with the our image model and text model.

\subsubsection{IIN}
The Interactive Inference Network (IIN) architecture is originally proposed by Gong et al. \cite{26} for TE. The IIN architecture is a typical relation-based entailment recognition architecture which models the interaction relation between each pair of the premise token and hypothesis token by building the interaction space. IIN is altered to apply to VE by using our image model and text model as embedding layer and encoding layer of IIN architecture. The interaction layer is implemented as element-wise product. The feature extraction layer is implemented as a plain ResNet and the last fully connected layer of ResNet is directly used as prediction output layer.

\subsection{Experiment Configurations}
The number of image features in each premise image is fixed to $m=36$ for both grid features and RoI features. The text is tokenized by spaCy\footnote{https://spacy.io/}  and the GloVe.840B.300d\footnote{http://nlp.stanford.edu/data/glove.840B.300d.zip} is used as word embedding which represents each token as a 300-d vector and is frozen during training.

For dataset SNLI-VE, as we can see in Table~\ref{table1}, Xie et al. provide the experiment result of TD/BUTD and the EVE model they proposed.

The experiment is implemented with PyTorch \cite{47}. Models are trained with cross entropy loss optimized by SGD with momentum 0.9 or Adam \cite{48}. The max training epoch is set to 100 and the batch size is 64. The learning rate is set as $1e-4$ by default. We apply a learning rate decay strategy that the learning rate is decayed by a coefficient 0.1 whenever the loss on validation set does not decrease for 2 epochs. The training dumps a checkpoint after an epoch and the checkpoint with best validation set accuracy is evaluated for test set accuracy.

For TD/BUTD, we re-implement the model following the existing implementation\footnote{https://github.com/claudiogreco/coling18-gte} and set major configurations according to Do et al. \cite{5}. For EVE, we implement the model and set major configurations according to the paper \cite{2} since no code has been released yet. The migrated rLSTM, IIN and our AlignVE are implemented with the same image model and text model. The configurations of AttEnc in them are tuned to 6 heads and 2 layers.

To control randomness and enhance experiment reproducibility, the random seeds of both Python and PyTorch are set to 12345 and the cuDNN deterministic mode is turned on.

\subsection{Results and Analyses}
\subsubsection{Comparative Experiment}

The experiment results on SNLI-VE dataset are shown in Table~\ref{table1}. The results reported by Xie et al. \cite{2} are also included for a better comparison.

We realize the fairness of experiments by controlling the variables of data processing and feature extraction sections, thus ensuring the experimental rigor. In experiment, the test accuracy of TD and BUTD are 71.04\% and 70.97\%, higher than Xie et al.’s work by 0.74\% and 2.07\% separately. The test accuracy of EVE-Image and EVE-RoI are 71.43\% and 70.43\% respectively. Compared with Xie et al.’s results, our EVE-Image is higher by 0.27\% and our EVE-RoI is lower by 0.04\%. In all, the re-evaluation experiments generally reproduce the results reported by Xie et al. It is acceptable to have some differences within a reasonable range because of the differences in data preprocessing, feature extraction, model and training configurations, randomness control and so on. Thus, our re-implementation of previous experiments meets the expectation and the accuracy is higher in general, indicating it is reasonable to use the reproduced experimental results as the benchmark model for the comparative experiment. 

The rLSTM and IIN are selected as typical examples of content-based and relation-based entailment recognition architectures respectively. With grid image features, the test accuracy of our migrated rLSTM and IIN are 71.25\% and 71.34\%, both are slightly lower than EVE-Image and higher than TD. With RoI image features, the test accuracy of rLSTM and IIN are 71.26\% and 71.30\%, both higher than those of BUTD and EVE-RoI. As for the migrated rLSTM and IIN, IIN performs better than rLSTM by a very slight margin.

Our AlignVE achieves 72.45\% with grid image features and 72.20\% with RoI image features. Compared to the results of models in previous VE researches, our AlignVE outperforms TD by 1.41\%, BUTD by 1.23\%, EVE-Image by 1.02\% and EVE-RoI by 1.77\%. Compared with the migrated IIN, our AlignVE has the same embedding layer and encoding layer, but the interaction layer, feature extraction layer and output layer are simpler. Specifically, our AlignVE models the relation between the premise-hypothesis pair as an alignment matrix rather than an interaction space. Additionally, the feature extraction layer and output layer of our AlignVE are simply implemented with a pooling and a fully connected layer instead of a deep CNN. Although the structure is simpler, the experiment results proves that AlignVE still outperforms migrated IIN by 1.11\% and 0.90\% with grid and RoI image features respectively.

The results of experiments on SNLI-VE-2.0 dataset is shown in Table~\ref{tab01}. Our re-implementation on this dataset verifies that the corrected dataset has no sigificant effect on the model performance by a drop of BUTD model from 70.97\% on SNLI-VE dataset to 69.26\% on SNLI-VE-2.0 dataset. Furthermore, our model AlignVE also shows that the improved dataset has no explicit difference in model performance with a slight increase of AlignVE model from 72.45\% to 72.67\%. Therefore, the SNLI-VE-2.0 dataset may not be meaningful from the current experiment data. Recall that only the validation and test set are corrected for data in \textit{neutral} class. The training set is needed to be corrected as well if a better model performance is required.

\subsubsection{Ablation Experiment}
We do the ablation experiment to show the effectiveness of the AttEnc applied in our study. To replace the AttEnc in the model, MLP is used in visual feature extraction module to achieve visual features and GRU \cite{45} is taken in text feature extraction module to obtain textual features. In this ablation experiment, we keep the other settings the same as described in Experiment Configurations and do the implementation based on the SNLI-VE dataset. As shown in Table~\ref{table2}, the test accuracy of various models all drop by 1\% approximately and the accuracy of the model AlignVE is still better than the accuracy of other compared models after the ablation, which proves AttEnc is effective and indicates that the proposed alignment-based VE architecture is meaningful as well.

What is more, to further demonstrate the model effectiveness, we replace the alignment matrix with a co-attention layer \cite{30}, where we take the text feature and image feature as inputs and send the output of the co-attention layer to the classification layer in the same way. This co-attention layer uses the parallel co-attention mechanism which generates image and text attention simultaneously. The parameters used are the same as what we use in AlignVE model and the dataset used is SNLI-VE. The results in Table~\ref{tableCo} show that the test accuracy of our model has 5.42\% and 6.44\% higher than replacing with a co-attention layer by using grid image features and RoI image features respectively, which indicates that the alignment matrix is more effective compared to the co-attention layer.

\subsubsection{ResNet Series Experiment}
In this paper, we also explore the influence of applying different pre-training models (i.e. ResNet series \cite{42}) on the AlignVE model (with grid image features) performance. The experiment results can be found in Table~\ref{tableResNet}. From the table, it can be concluded that the performance of the AlignVE model with grid image features becomes higher when the feature dimensionality becomes larger. Meanwhile, with the feature dimensionality fixed, the performance of the model increases at first as the layer number increases and then decreases. The possible reason is that the image dataset is relatively simple and small to make the features easy to identify. So the ResNet-101 is a better fit than ResNet-152, which proves that deeper is not necessarily the better for deep learning models as well.

\subsubsection{Visualization Analysis}
Our AlignVE model can also support the visualization analysis since the alignment matrix is 2-dimensional which can be expressed in the form of visual attention. Thus, by applying the alignment matrix into visualization, we can directly obtain and examine the cross-modal alignment between the premise and hypothesis the model has learned.

Considering each row of the alignment matrix corresponds to a section of a premise image and each column corresponds to a hypothesis word, we do the visualization analysis in the unit of column. The alignment value of every column is counted as the visual attention of each word and is painted on the premise image with the jet color mapping. In this case, the alignment can be seen through the mask rendering. The more red in the image means the stronger alignment and the blue indicates the weak alignment relation. 

In Figure~\ref{fig2}, it shows a picture with three different types of hypothesises which correspond to the three classification result \textit{neutral}, \textit{entailment} and \textit{contradiction}. The semantic of the premise image is that a man is about to go out of a swimming pool. The first hypothesis is ``A man enjoys his vacation". It can be seen from this case that, the visual attention of ``man" focuses on the middle of the image which is the area that indicates the man. Since the words ``enjoys" and ``vacation" are hardly to conclude from this image, the most of the scene corresponding to these two words are blue and green. Therefore, this case demonstrates the model's judgement ability when there is no evidence to prove the hypothesis matches the premise image. The second text hypothesis is ``The man is all wet". From the visualization results, it can be discovered that the middle of the image is colored in red when the words come to ``man" and ``wet".  Therefore, the alignment results imply that the model is able to learn the alignments of 
the figure and environment between premise and hypothesis. Also, the model's understanding is corresponding to the human intuition. In the third case, the hypothesis is ``The man was laying by the pool".  From the visualization, the words ``man" and ``pool" correspond to the person and the swimming pool in the image, while the visual attention of words ``laying" and ``by" shows the weak alignment relation to the person's action and the environment. In summary, the model has the capability of understanding the \textit{contradiction} classification given a pair of a premise image and a hypothesis text.

\section{Conclusion}
In this paper, we propose a new alignment-based visual entailment (AlignVE) model. With the intuition that entailment recognition tasks like VE should focus on understanding the relation between the premise and hypothesis, the AlignVE calculates the relation of the premise-hypothesis pair as an alignment matrix and recognizes VE based on it. Experiments show that AlignVE outperforms the previous VE models and the migrated typical TE models on SNLI-VE dataset in the same experiment settings, indicating that AlignVE architecture is simple and powerful.

\bibliographystyle{IEEEtran}
\bibliography{IEEEabrv, IEEEexample, B}

\begin{IEEEbiography} [{\includegraphics[width=1in,height=1.25in,clip,keepaspectratio]{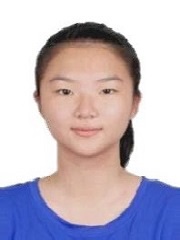}}]{Biwei Cao}
received the B.S. degree in Software Engineering from Australian National University, in 2019, the M.S. degree in Computing from Australian National University, in 2020. She is now working toward the Ph.D degree from the School of Cyber Science and Engineering, Southeast, University, Nanjing, China, Her research interests include affective computing, gesture recognition, social computing and language generation.
\end{IEEEbiography}
% \vfill
\begin{IEEEbiography} [{\includegraphics[width=1in,height=1.25in,clip,keepaspectratio]{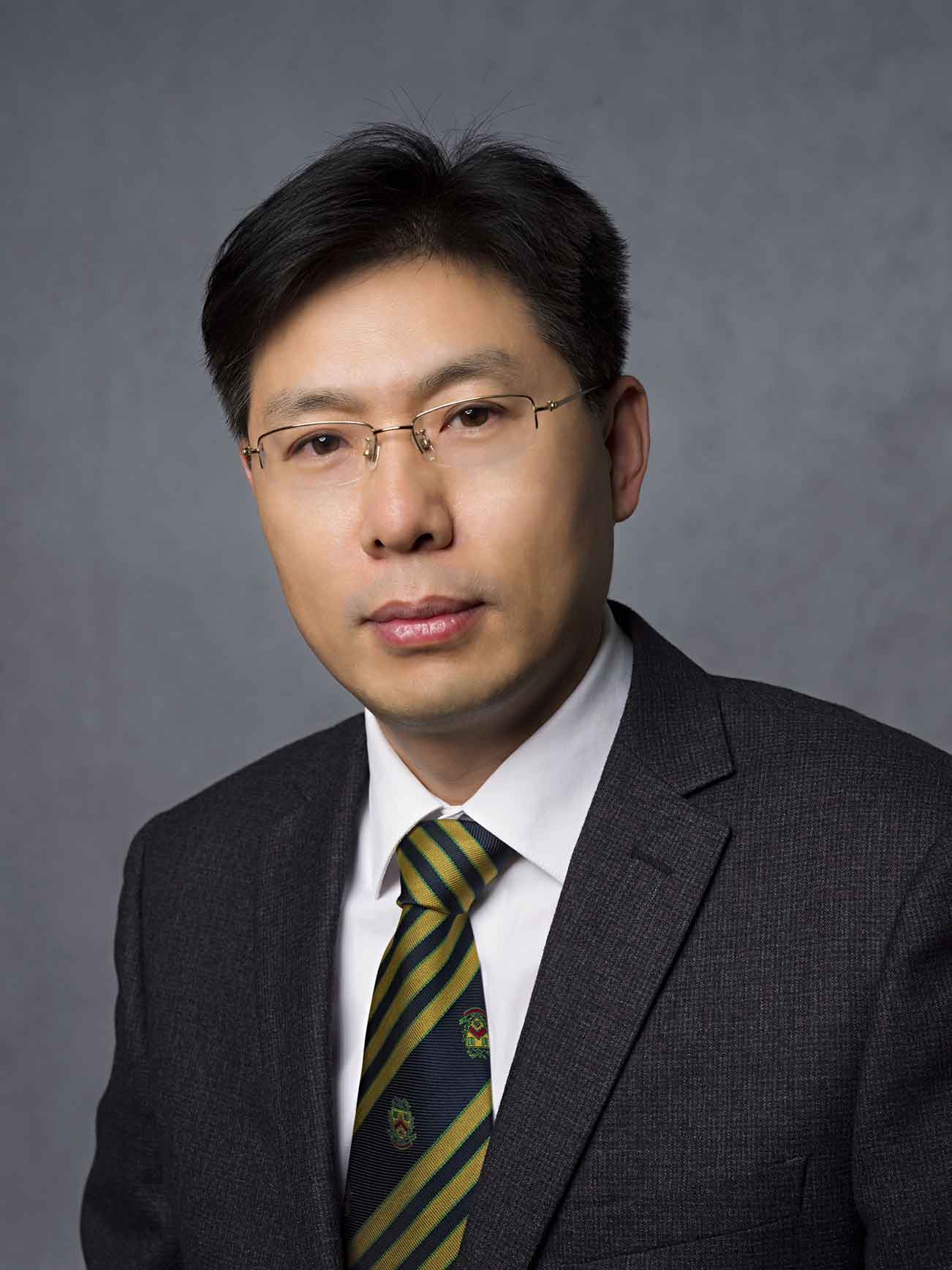}}]{Jiuxin Cao}
received the Ph.D degree from Xi’an Jiaotong University in 2003. He is currently a professor at the School of Cyber Science and Engineering in Southeast University, Nanjing, China. His research interests include computer networks, social computing, affective computing, behavior analysis, and big data security and privacy preservation.
He is the Director of Jiangsu Provincial Key Laboratory of Computer Network Technology, the Senior Member of China Computer Federation, the Member of Chinese Information Processing Society of China, the Fellow of Jiangsu Computer Society, the Member of Jiangsu Information Security Standardization Committee, and the Member of JSAI-ISA.
\end{IEEEbiography}

\begin{IEEEbiography} [{\includegraphics[width=1in,height=1.25in,clip,keepaspectratio]{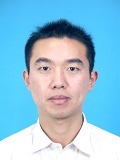}}]{Jie Gui}
 received the M.S. degree in computer applied technology from Hefei Institutes of Physical Science, Chinese Academy of Sciences, Hefei, China, in 2007, and the Ph.D. degree in pattern recognition and intelligent systems from the University of Science and Technology of China, Hefei, China, in 2010. He is currently a Professor in the School of Cyber Science and Engineering, Southeast University. He has published more than 40 papers in international journals and conferences such as IEEE TPAMI, IEEE TNNLS, IEEE TCYB, IEEE TIP, IEEE TCSVT, IEEE TSMCS, KDD, and ACM MM. He is the Area Chair, Senior PC member, or PC Member of many conferences such as NeurIPS and ICML. His research interests include machine learning, pattern recognition, and image processing.
\end{IEEEbiography}
\begin{IEEEbiography} [{\includegraphics[width=1in,height=1.25in,clip,keepaspectratio]{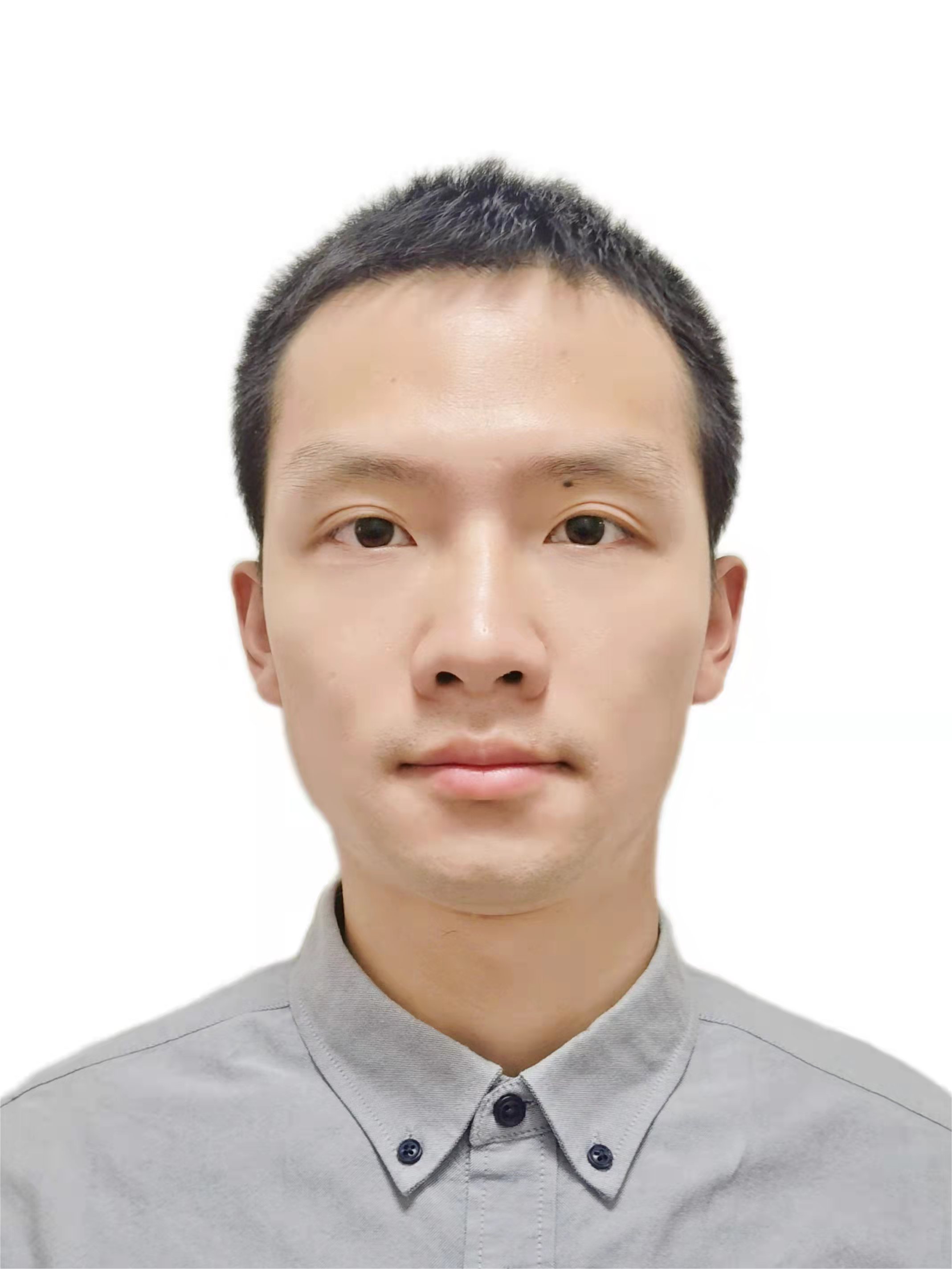}}]{Jiayun Shen} received the M.S. degree from the School of Cyber Science and Engineering, Southeast University, in 2020. His research interests includes object detection, visual analysis, and cross modal analysis.
\end{IEEEbiography}
\begin{IEEEbiography} [{\includegraphics[width=1in,height=1.25in,clip,keepaspectratio]{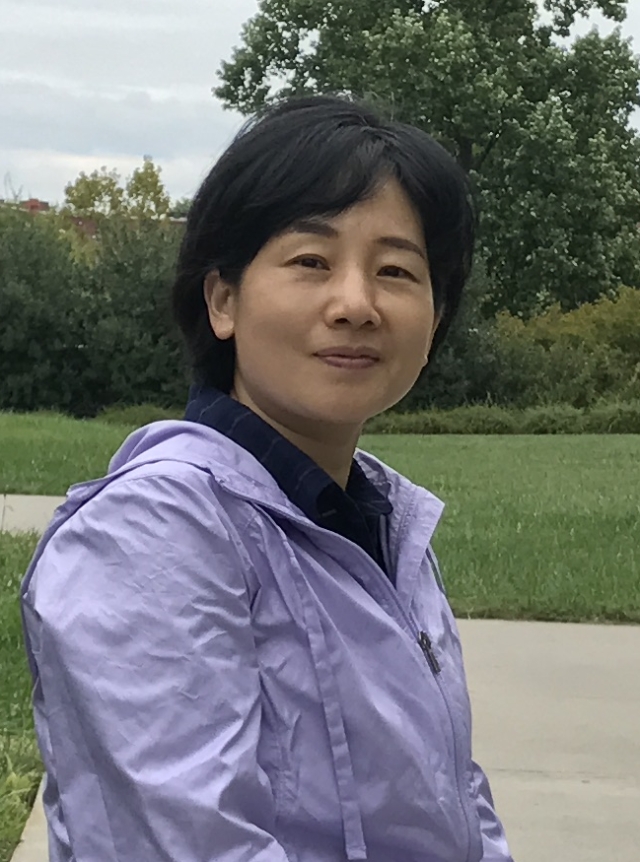}}]{Bo Liu}
 works as a professor and the doctoral advisor with Southeast University, China. She received her
doctoral degree from Southeast University. She won the first class Science and Technology Progress Award of MoE in 2009, and she is currently working on two NSF projects. She has published more than 60 papers and most of them have been published in reputed journals and conferences including WWW, WWWJ, ToN and et al. Her current main research interests include spammer detection in social network, the evolution of social community, social influence, and social recommendation.
\end{IEEEbiography}
\begin{IEEEbiography} [{\includegraphics[width=1in,height=1.25in,clip,keepaspectratio]{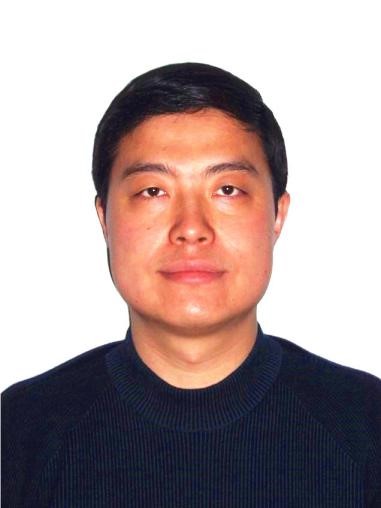}}]{Lei He} received the Ph.D. degree from Information Engineering University, China, in 2008. He is an Associate Researcher of Information Engineering University, China. His current research interests include Cyberspace security and Mimic defense.
\end{IEEEbiography}

\begin{IEEEbiography} [{\includegraphics[width=1in,height=1.25in,clip,keepaspectratio]{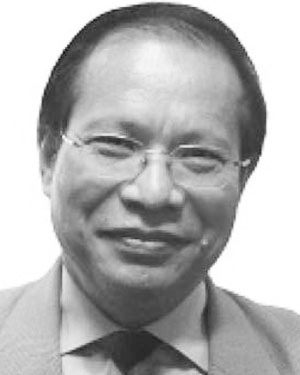}}]{Yuan Yan Tang}
is an IEEE Life Fellow, IAPR Fellow, and AAIA Fellow. He currently is the Director of Smart City Research Center in Zhuhai UM Science \& Technology Research Institute, is also the Emeritus Chair Professor at University of Macau and Hong Kong Baptist University, Adjunct Professor at Concordia University, Canada. His current research interests include artificial intelligence, wavelets, pattern recognition, and image processing. He has published more than 600 academic papers and is the author (or coauthor) of over 25 monographs, books and bookchapters. He is the Founder and Editor-in-Chief of SCI journal “International Journal on Wavelets, Multiresolution, and Information Processing (IJWMIP)”. Dr. Tang is the Founder and General Chair of the series International Conferences on Wavelets Analysis and Pattern Recognition (ICWAPRs). He is the Founder and Chair of the Macau Branch of International Associate of Pattern Recognition (IAPR). He has serviced as general chair, program chair, and committee member for many international conferences. Dr. Tang served as the Chairman of 18th ICPR, which is the first time that the ICPR was hosted in China.
\end{IEEEbiography}

% \vfill
% \newpage
\begin{IEEEbiography} [{\includegraphics[width=1in,height=1.25in,clip,keepaspectratio]{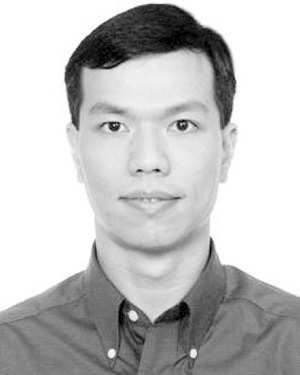}}]{James Tin-Yau Kwok}
 received the Ph.D. degree in computer science from The Hong Kong University of Science and Technology, Hong Kong, in 1996. He is currently a Professor with the Department of Computer Science and Engineering, The Hong Kong University of Science and Technology. His current research interests include kernel methods, machine learning, pattern recognition, and artiﬁcial neural networks. He received the IEEE Outstanding Paper Award in 2004 and the Second Class Award in Natural Sciences from the Ministry of Education, China, in 2008. He has been a Program Co-Chair for a number of international conferences, and served as an Associate Editor for the IEEE TRANS-ACTIONS ON NEURAL NETWORKS AND LEARNING SYSTEMS from 2006 to 2012. He is currently an Associate Editor of \textit{Neurocomputing}.
\end{IEEEbiography}
\vfill

\vfill

\end{document}